\documentclass[conference]{IEEEtran}
\IEEEoverridecommandlockouts
\usepackage{cite}
\usepackage{amsmath,amssymb,amsfonts}
\usepackage{algorithmic}
\usepackage{graphicx}
\usepackage{textcomp}
\usepackage{xcolor}
\usepackage{subcaption}
\def\BibTeX{{\rm B\kern-.05em{\sc i\kern-.025em b}\kern-.08em
    T\kern-.1667em\lower.7ex\hbox{E}\kern-.125emX}}

\usepackage{cite}
\usepackage{amsmath,amssymb,amsfonts}
\usepackage{algorithmic}
\usepackage{graphicx}
\usepackage{textcomp}
\usepackage{xcolor}
\usepackage{subcaption} %
\usepackage{graphicx} %
\usepackage[numbers]{natbib}  %
\usepackage{caption} %
\usepackage{algorithm}
\usepackage{listings}
\usepackage{algorithmic}
\usepackage{amsmath}
\usepackage{booktabs}
\usepackage{multirow}
\usepackage[outdir=./]{epstopdf}
\usepackage{enumitem}
\usepackage{caption}
\usepackage{subcaption}
\usepackage{newfloat}
\usepackage{graphicx}
\usepackage{svg} %
\usepackage[normalem]{ulem}
\usepackage{framed}
\usepackage{mdframed}
\usepackage{xcolor}
\usepackage{lipsum}
\usepackage{float}
\usepackage{hyperref}

\definecolor{shadecolor}{gray}{0.9}

\newlist{todolist}{itemize}{2}
\setlist[todolist]{label=$\square$}
\usepackage{pifont}

\def\BibTeX{{\rm B\kern-.05em{\sc i\kern-.025em b}\kern-.08em
    T\kern-.1667em\lower.7ex\hbox{E}\kern-.125emX}}

\begin{document}

\title{TaxAgent: How Large Language Model Designs Fiscal Policy\\
\thanks{*Corresponding author: Yongfeng Huang (1155187959@link.cuhk.edu.hk)}
}

\author{\IEEEauthorblockN{1\textsuperscript{st} Jizhou Wang}
\IEEEauthorblockA{\textit{Illinois Institute of Technology} \\
Chicago, United States\\
jwang307@hawk.iit.edu}
\and
\IEEEauthorblockN{1\textsuperscript{st} Xiaodan Fang}
\IEEEauthorblockA{\textit{University of Texas at Austin} \\
Austin, United States \\
denisfang@utexas.edu}
\and
\IEEEauthorblockN{2\textsuperscript{nd} Lei Huang}
\IEEEauthorblockA{\textit{Jiangxi Normal University} \\
Nanchang, China \\
1149505181@qq.com}
\and
\IEEEauthorblockN{3\textsuperscript{rd} Yongfeng Huang*}
\IEEEauthorblockA{\textit{Chinese University of Hong Kong} \\
Hong Kong, China\\
1155187959@link.cuhk.edu.hk}
}
\maketitle

\begin{abstract}
Economic inequality is a global challenge, intensifying disparities in education, healthcare, and social stability. Traditional systems like the U.S. federal income tax reduce inequality but lack adaptability. Although models like the Saez Optimal Taxation adjust dynamically, they fail to address taxpayer heterogeneity and irrational behavior. This study introduces TaxAgent, a novel integration of large language models (LLMs) with agent-based modeling (ABM) to design adaptive tax policies. In our macroeconomic simulation, heterogeneous H-Agents (households) simulate real-world taxpayer behaviors while the TaxAgent (government) utilizes LLMs to iteratively optimize tax rates, balancing equity and productivity. Benchmarked against Saez Optimal Taxation, U.S. federal income taxes, and free markets, TaxAgent achieves superior equity-efficiency trade-offs. This research offers a novel taxation solution and a scalable, data-driven framework for fiscal policy evaluation.
\end{abstract}

\begin{IEEEkeywords}
Large language models (LLMs), Adaptive tax policies, Macroeconomic simulation
\end{IEEEkeywords}

\section{Introduction}
\label{sec:intro}

Economic inequality is a critical global issue with profound social, political, and economic impacts. Research highlights its detrimental effects on education, healthcare, political stability, and economic growth\cite{FLUG1998465, 10.1093/isr/viab058, GLAESER2003199}. Tackling inequality is essential for building a fair and prosperous society.

Progressive taxation emerged to address inequality. By imposing higher tax rates on higher incomes, systems like the US federal income tax have shown potential to reduce poverty and improve health, education, and economic opportunities\cite{NBERw21340}\cite{NBERw21211}. However, these systems are static, limited by legislative constraints and unable to adapt to changing economic conditions.

Research on dynamic optimal taxation has evolved over time. The Mirrlees framework introduced incentive compatibility\cite{10.2307/2296779}, while Diamond and Mirrlees\cite{RePEc:aea:aecrev:v:61:y:1971:i:1:p:8-27} emphasized production efficiency. Atkinson and Stiglitz\cite{ATKINSON197655} advocated for income-based taxation to simplify tax structures. Saez\cite{10.1111/1467-937X.00166} developed a rule-based framework, optimizing social welfare functions with dynamic tax adjustments. Despite advancements, current dynamic taxation theories face two significant challenges.

\begin{itemize}

\item \textbf{Calculations of economic metrics such as elasticity and social welfare lack validation.}

\item \textbf{Limited capacity to account for the behavioral heterogeneity and irrationality of taxpayers.}

\end{itemize}

Agent-based modeling (ABM) and large language models (LLMs) offer innovative solutions to challenges in tax policy optimization. Replacing rule-based agents with LLM-based ones allows adaptive tax optimization by simulating realistic taxpayer behavior and eliminating rigid assumptions. LLMs enhance policy design through advanced reasoning and data interpretation\cite{li2024econagentlargelanguagemodelempowered}, without reliance on contentious welfare calculations.

This study introduced a taxation evaluation framework, comprising three components: the \textbf{TaxAgent} (government), H-Agents Group (heterogeneous households), and a macroeconomic simulation environment. Iterative interactions between households and the government model the dynamic effects of tax policies. The TaxAgent, powered by an LLM, analyzes economic metrics and household behavior to propose adaptive tax rates aligned with societal goals. H-Agents Group represents diverse households, making decisions about work and consumption based on economic conditions and learned experiences. The macroeconomic environment processes these interactions, updating metrics like production, wages, and prices, enabling experimentation with tax policies.

The TaxAgent addresses limitations of traditional models by leveraging LLMs to synthesize data and trends, overcoming rigid assumptions on welfare calculation. The H-Agents Group simulates human-like decision-making, capturing heterogeneous and irrational household behaviors, thereby providing realistic responses to tax policies.

The performance of the TaxAgent is benchmarked against the Saez Optimal Taxation, the U.S. Federal Income Tax system, and a free-market scenario. Results highlight its transformative potential. The TaxAgent outperforms traditional systems in balancing equality and productivity, demonstrating robustness across LLM implementations. 

This study makes the following contributions:
\begin{itemize}
\item \textbf {A Scalable and Cost-Effective Policy Evaluation Framework:} We introduce a simulation environment that integrates simulated policy makers, market dynamics, and heterogeneous households, overcoming the limitations of traditional economic models, empowering real-world policymakers to test and refine policies in a controlled setting, greatly reducing the cost of policy evaluation.
\item \textbf {Innovative Use of LLMs in Tax Policy Design:} By employing LLMs as tax planners and taxpayers, we eliminate the need for rigid assumptions, enabling dynamic, adaptive, and data-driven approaches to taxation.
\item \textbf {Empirical Validation:} The TaxAgent demonstrates practical superiority over established systems, showcasing its ability to effectively solve the equity-efficiency dilemma. The TaxAgent exemplifies how AI-driven tools can pioneer new frontiers in fiscal policy, offering adaptive and equitable solutions to pressing global issues.
\end{itemize}
\section{Related Work}
\subsection{Traditional Tax Systems}
Progressive taxation is relatively simple, imposing higher tax rate on higher income. Empirical studies proved its effectiveness in ameliorating economic inequality~\cite{NBERw21340, NBERw21211} but researchers also pointed out that it lacks adaptability to dynamic economic conditions~\cite{Foo2019ProcessAC, Patjoshi2015DesignAD}.

Optimal taxation explores tax systems that maximize social welfare while accounting for economic constraints and behavioral responses\cite{10.1257/jep.25.4.165}. It can be adjusted according to dynamic economic conditions. Modern frameworks were pioneered by Mirrlees\cite{10.2307/2296779} and Diamond and Mirrlees\cite{RePEc:aea:aecrev:v:61:y:1971:i:1:p:8-27}, aiming to maximize aggregate utility. Saez is one of the largest contributors to this basis. Saez\cite{10.1111/1467-937X.00166} derived optimal nonlinear tax rates by modeling earnings elasticity and income distribution. Diamond and Saez\cite{10.1257/jep.25.4.165} extended this framework, focusing on maximizing social welfare while mitigating income inequality, constructing a closed-loop optimal taxation system.

Economists emphasized the importance of the behavioral responses of taxpayers. Piketty, Saez, and Stantcheva \cite{10.1257/pol.6.1.230} explored the elasticity of the top tax rates and their influence on labor supply and tax avoidance, highlighting the importance of behavioral responses in tax policy design. Kroft, Kucko, Lehmann, and Schmieder \cite{10.1257/pol.20180033} examined how unemployment and wage responses impact tax structures, advocating for the Earned Income Tax Credit (EITC) as a tool to support low-income households while maintaining working incentives.

\subsection{Artificial Intelligence(AI) in Economic Policy Research}
AI offers innovative tools for analyzing and optimizing macroeconomic policies, addressing limitations of traditional models, which rely on equilibrium assumptions. Reinforcement learning and Bayesian Neural Networks enable adaptive simulations and uncertainty quantification. For example, “The AI Economist” framework uses RL to co-adapt agents and social planners\cite{zheng2020aieconomistimprovingequality}. Integration with causal inference techniques further improves policy-impact assessment\cite{NBERc14009}.

ABM captures decentralized decision-making and complex phenomena like systemic risk\cite{AxtellFarmer2022}. ABMs are used to study business cycles, policy interventions, and inflation\cite{DelliGatti2018}. Enhanced computational techniques and high-quality data have improved their empirical validity, enabling applications such as tax policy optimization\cite{zheng2020aieconomistimprovingequality}.

Large Language Models (LLMs) introduce advanced reasoning capabilities to various subjects, including economic research, enabling market behavior simulation and policy evaluation~\cite{shen2025phyxdoesmodelwits,zhao2024competeaiunderstandingcompetitiondynamics, nie2024surveylargelanguagemodels}.

Existing work forms a rule-based framework for optimal taxation and recognizes the impact of taxpayer heterogeneity on optimal taxation design. Nevertheless, current optimal taxation makes rational-man supposition and oversimplified social welfare calculations. In this work, we integrated the advancement in ABMs and LLMs, replaced predetermined rules with LLM-based agents, simulated human-like policy responses, and dynamically adjusted tax rates to generate the optimal social outcome.
\section{Taxation Evaluation Framework}
Our framework integrates three core components—the TaxAgent (government), H-Agents Group (households), and the macroeconomic simulation environment—into a policy evaluation framework that models household-government interactions within an evolving economy. The system operates as follows:
\begin{itemize}
    \item \textbf{Household Decision-Making:} H-Agents, representing households, observe economic dynamics such as taxation and market conditions from the macroeconomic environment while incorporating past experiences. They decide on work and consumption propensities based on these inputs.
    \item \textbf{Macroeconomic Environment Dynamics:} Heterogeneous household decisions are processed, updating metrics like production, wages, and prices, reflecting supply-demand dynamics and financial market influences.
    \item \textbf{Government Decision-Making via the TaxAgent:} The TaxAgent, representing the government, analyzes updated economic metrics and household behavior using an LLM. It proposes tax rates optimized for social goals.
    \item \textbf{Iterative Feedback:} The TaxAgent’s new rates are implemented, and updated metrics feed back to H-Agents for further decision-making. This cycle continues for a set number of iterations.
\end{itemize}
The following subsections detail the roles and mechanisms of the three core components. Specific metric calculations and prompts for agents are provided in the appendix.

\begin{figure*}[t]
\centering
    \includegraphics[width=0.75\linewidth]{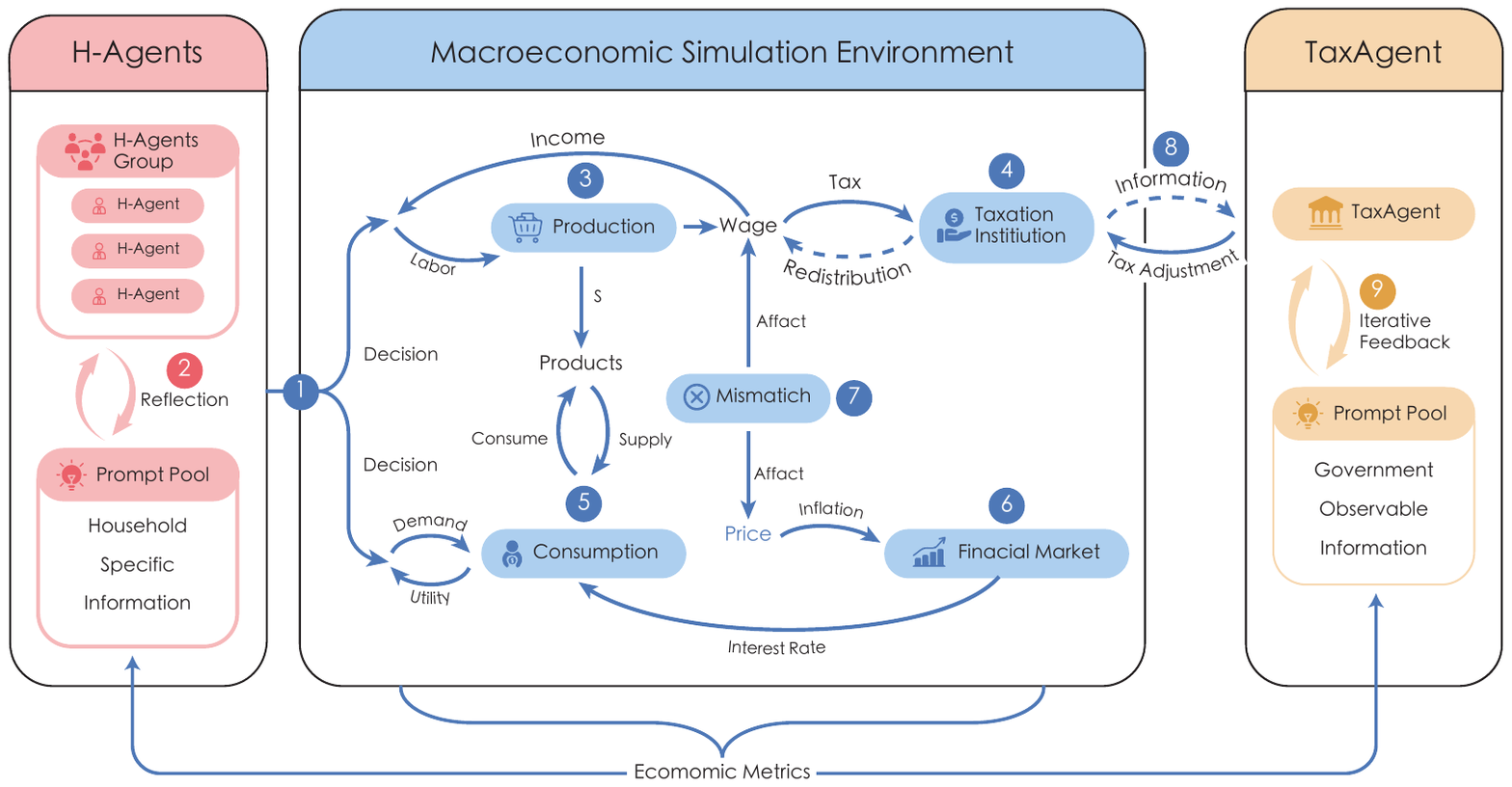}
\caption{The illustration of the Taxation Evaluation System.}
\label{fig:framework}
\end{figure*}

\subsection{H-Agent: Agent as a Household}
One H-Agent represents one household, using an LLM to model decision-making in response to economic conditions. By interacting with the macroeconomic environment and the TaxAgent, H-Agents Group collectively influence macroeconomic outcomes and societal impacts of tax policies.

Each H-Agent operates through two modules:
\begin{itemize}
    \item \textbf{Decision-Making:} H-Agents determine production and consumption propensities based on economic inputs, including taxation, labor markets, price levels, and financial markets.
    \item \textbf{Self-Reflection:} A reflection module enhances decision-making by maintaining a memory pool of past economic data and decisions. Quarterly, H-Agents refine future behavior by reviewing this information.
\end{itemize}

\subsubsection{Decision-Making Process}
H-Agents decide on working and consumption propensities ($p^{w}_{i}, p^{c}_{i}$) based on current economic observations and past reflections, as shown in Figure~\ref{fig:framework}, process \normalsize{\textcircled{\scriptsize{1}}}\normalsize:
\begin{equation}
(p^{w}_{i}, p^{c}_{i}) = {H}_{i}(Pmt_i, \theta^{R}_{i})
\end{equation}
Here, ${H}_{i}$ is the decision function for the $i$-th household, $Pmt_i$ includes observable economic data, and $\theta^{R}_{i}$ represents reflection-based parameters from previous decisions.

\subsubsection{Self-Reflection}
At the end of each quarter, as loop \normalsize{\textcircled{\scriptsize{2}}}\normalsize
~in Figure~\ref{fig:framework}, H-Agent reviews its decision and economic history to update its reflection parameters:
\begin{equation}
\theta^{R}_{i} \leftarrow \text{H}_{i}(\text{Memo}_i)
\end{equation}
where $\text{Memo}_i$ represents the $i$-th household’s prior economic prompts and decision history.

\subsection{TaxAgent: Agent as a Government}
The TaxAgent is the central authority of the macroeconomic simulation, leveraging an LLM to dynamically adjust tax rates, balancing two societal goals: productivity and equality.

The TaxAgent iteratively performs two core steps:
\begin{itemize}
    \item \textbf{Tax rate adjustment:} Using a heuristic prompt that combines household data, global performance metrics, and decision-making flexibility, the TaxAgent integrates traditional tax data with the LLM’s reasoning. It analyzes economic trends and the productivity-equality trade-off to generate tax rates aimed at optimizing societal objectives.
    \item \textbf{Iterative Feedback:} Generated tax rates influence household behavior and economic conditions within the simulation. Updated metrics are fed back to the TaxAgent, enabling continuous refinement of its strategy through an iterative feedback loop.
\end{itemize}

\subsubsection{Tax Rate Adjustment}
The tax rate adjustment process corresponds to loop \normalsize{\textcircled{\scriptsize{8}}}\normalsize ~in Figure~\ref{fig:framework} and can be represented as follows:
\begin{equation}
TX = {Gov}(Pmt, \theta_{G}, \theta_{H})
\end{equation}
where $Pmt$ includes household data, global performance metrics, and decision-making flexibility; $\theta_{G}$ represents the LLM’s trained parameters on the government’s optimal taxation strategy and $\theta_{H}$ represents the trained parameters on household reactions to adjusted tax rate.

\subsubsection{Iterative Feedback Mechanism}
Once proposed, the tax rates are implemented in the simulation environment, influencing household behavior and market dynamics. The resulting economic metrics are then fed back to the TaxAgent, forming an iterative feedback loop as shown in Figure~\ref{fig:framework}, loop \normalsize{\textcircled{\scriptsize{9}}}\normalsize :
\begin{equation}
\theta_{G}, \theta_{H} \leftarrow {Gov}(\text{Pmt}_{upd}, \theta_{G}, \theta_{H})
\end{equation}
Where $\text{Pmt}_{upd}$ includes latest household data and global performance metrics.

\subsection{Macroeconomic Simulation Environment}
The macroeconomic simulation environment models key aspects of a real-world economy. It includes four modules: production, taxation, consumption, and the financial market, which interact dynamically through feedback loops.

Economic metrics in each module are updated based on decisions of the TaxAgent and H-Agents Group. These metrics, in turn, inform agent decisions, creating a cyclical feedback loop that begins with production and wage distribution, followed by taxation, income allocation, and adjustments to wages and prices based on production-consumption dynamics.

\subsubsection{Production Module}
Production constitutes the starting point of economic activity, as shown in Figure~\ref{fig:framework}, Module~\normalsize{\textcircled{\scriptsize{3}}}\normalsize. The production is determined by the total labor supplied:
\begin{equation}
S = \sum_{j=1}^{N} s_j
\end{equation}
The inventory $G$ is updated after production as follows:
\begin{equation}
G \leftarrow G + S
\end{equation}
Households receive wages upon completion of production. However, wage determination is discussed later in this section and in the appendix for its complexity.

\subsubsection{Taxation Module}
Taxation, central to this study, follows wage distribution. The taxation is bracketed: income in each bracket is taxed at a specific rate. In addition, redistribution is even and implicit, emulating real-world scenarios:
\begin{equation}
    \ z_i = z_i^{pre} - T(z_i) + z^r
\end{equation}
Where $z_i$ is individual income, $z_i^{pre}$is pre-tax income, $T(z_i)$ is tax levied, and $z^r$ is the redistribution.

\subsubsection{Consumption Module}
After taxation, households allocate post-tax income between consumption and savings. Total demand is the sum of individual demands, and inventory $G$ updates dynamically:
\begin{equation}
    D = \sum_{j=1}^{N} d_j
\end{equation}
\begin{equation}
    G \leftarrow G - d_j
\end{equation}
Please refer to the appendix for more details.

\subsubsection{Financial Module}
The financial market incorporates the interest rate, a critical metric influencing household savings, corporate borrowing, and government policy. Savings increase annually by the prevailing interest rate $r$:
\begin{equation}
    s_i \leftarrow s_i\times (1 + r).
\end{equation}
Interest rates are dynamically adjusted based on the unemployment rate and inflation rate ~\cite{dawid2018agent}.

\subsubsection{Global Interdependency}
Interactions among production, taxation, consumption, and the financial market drive wage and price changes. When supply exceeds demand, prices drop, restraining profits and wages. Conversely, when demand exceeds supply, the opposite occurs. Please refer to the appendix for more details.

\subsection{Advancing Beyond Rule-Based Optimal Tax Systems}
The TaxAgent addresses key limitations of traditional optimal taxation.
\begin{itemize}
\item \textbf{Beyond Assumptive Optimization:} By combining data with an understanding of collective human welfare, the TaxAgent shifts from rigid optimization to adaptive decision-making, reflecting diverse societal perspectives as a "superposition of social consciousness."
\item \textbf{Modeling Irrational Behavior:} Learning from the irrational behaviors of H-Agents in the macroeconomic environment, the TaxAgent moves beyond the rational-agent assumption, producing superior outcomes.
\end{itemize}
\section{Experiments}
This section assesses the TaxAgent's ability to achieve balanced social outcomes compared to traditional tax systems, guided by the following research questions:
\begin{itemize}
    \item \textbf{RQ1}: How effectively does the TaxAgent balance equality and productivity compared to traditional tax systems?
    \item \textbf{RQ2}: What mechanisms enable the best-performing system to achieve superior results?
    \item \textbf{RQ3}: What are the macroeconomic side effects of deploying the TaxAgent?
\end{itemize}

\subsection{Baseline Tax Systems}
The TaxAgent's performance is benchmarked against three baseline tax systems:
\begin{itemize}
    \item \textbf{Saez Optimal Taxation}: It refines optimal income tax theory using  elasticities and income distributions, aiming to balance equality and efficiency~\cite{10.1111/1467-937X.00166}.
    \item \textbf{US Federal Income Tax}: It employs a progressive structure where higher-income households pay a larger share of taxes. This approach seeks to reduce inequality and benefit low-income groups.
    \item \textbf{Free Market}: In the free market, households retain all income without taxation or redistribution.
\end{itemize}
Please refer to the appendix for more details.

\subsection{Implementation Details}
The simulation comprises $N = 50$ households over $P = 120$ months. Four tax systems are tested: free market, U.S. federal income taxation, Saez optimal taxation, and the TaxAgent. Productivity is fixed at 1, with results remaining consistent across parameter variations. The simulation utilizes the qwen-turbo-2024-09-19 model via the OpenAI API. Detailed parameters and replicability information are provided in the appendix.

\subsection{Performance Evaluation Metrics}
Tax system performances are evaluated using equality and productivity per capita. Equality is calculated as the complement of the normalized Gini index, while productivity is defined as the current average wealth of H-Agents. The social outcome is assessed as the product of equality and productivity (the higher the better). Detailed calculations are included in the appendix.

\subsection{Experiment Results}
\subsubsection{The TaxAgent in Generating Social Outcomes (RQ1)}
\begin{figure}
    \centering
    \includegraphics[width=1\linewidth]{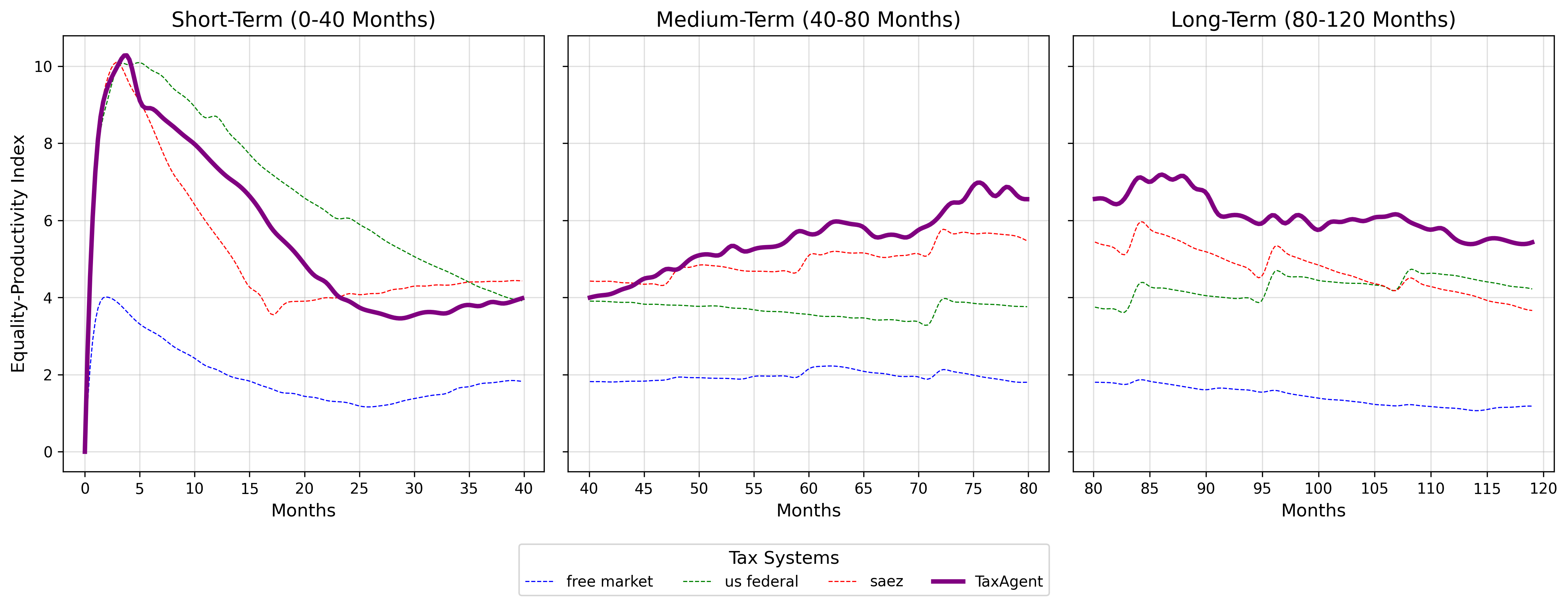}
    \caption{The social outcomes of all tax systems over 120 months. The TaxAgent (purple) performs significantly better in the long-term.}
    \label{fig:main exp}
\end{figure}

The TaxAgent achieves the best long-term social outcomes. Figure~\ref{fig:main exp} shows the Equality-Productivity Index in three baseline systems and the TaxAgent. Key findings include:
\begin{itemize}
    \item Short-term (0–40 months):The TaxAgent performs similarly to the Saez and US federal systems, with no clear advantage.
    \item Medium-term (40–80 months): The TaxAgent improves steadily, surpassing the Saez and US systems in balancing equality and productivity.
    \item Long-term (80–120 months): The TaxAgent maintains its lead, consistently outperforming other systems.
\end{itemize}
In summary, the TaxAgent demonstrates the best overall performance, while the free market performs the worst across all periods. The US federal and Saez systems show comparable, moderate results, with neither gaining a clear advantage.

The performance of the TaxAgent is robust. We implemented the TaxAgent using two different LLMs. It achieves superior results regardless of the underlying LLM. This indicates that the TaxAgent has low sensitivity to changes in its LLM base, enhancing its reliability. The results are shown in the Appendix.

\subsubsection{Mechanisms Behind TaxAgent Performance (RQ2)}
\begin{figure}
    \centering
    \includegraphics[width=1\linewidth]{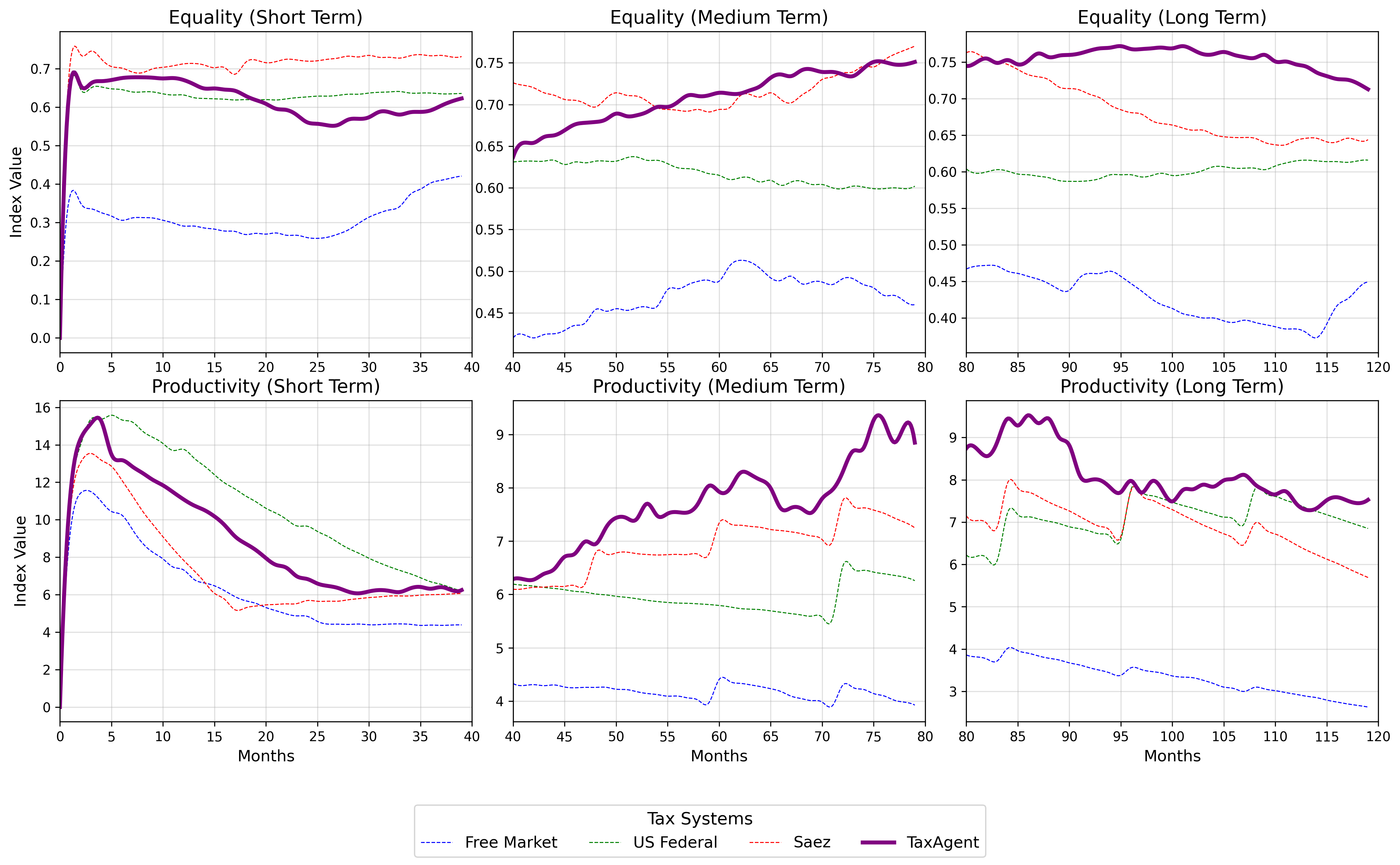}
    \caption{The equality(above) and productivity(below) performance of the TaxAgent. The TaxAgent demonstrates its prioritization on equality and its flexibility in making equality-productivity trade-offs.}
    \label{fig:e_q_separated}
\end{figure}

The TaxAgent achieves superior performance by prioritizing equality with dynamic flexibility.
\begin{itemize}
    \item \textbf{High Equality:} As shown in Figure~\ref{fig:e_q_separated}, the TaxAgent prioritizes equality, with values stabilizing between 0.6 and 0.75. While equality drops slightly in the short term, equality begins recovering near the 25th month.
    \item \textbf{Dynamic Flexibility:}  The TaxAgent adjusts tax rates dynamically. For instance, it sacrifices equality during sharp productivity declines (e.g., post-90th month) but strengthens equality when production stabilizes (e.g., around the 25th month). This adaptability drives its long-term success.
\end{itemize}

\begin{figure}
    \centering
    \includegraphics[width=1\linewidth]{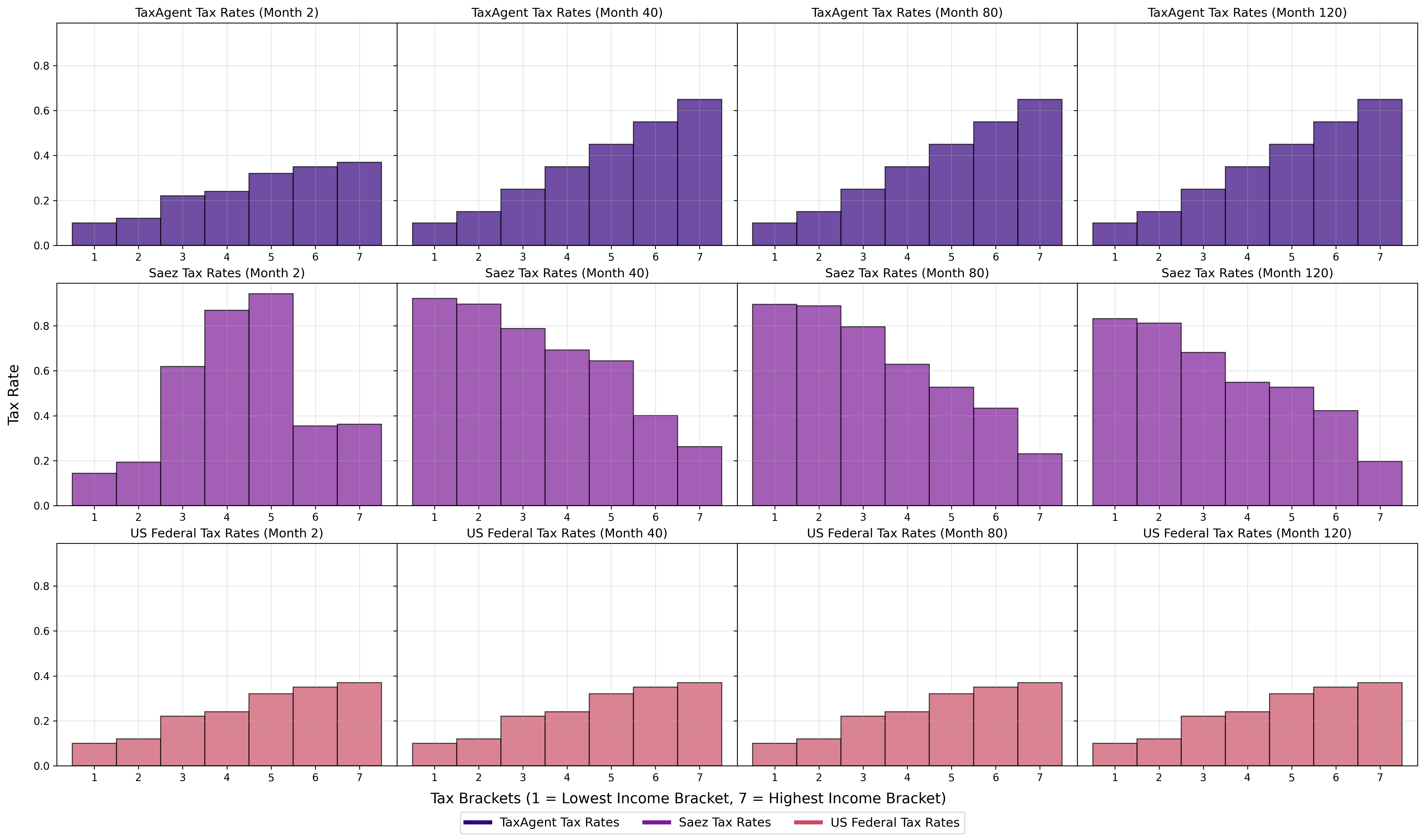}
    \caption{Sample tax rates for seven income brackets of the TaxAgent(top), the Saez taxation(middle), US federal income tax(bottom) over 120 months. Regressiveness of Saez taxation and rigidness of US federal income tax limit their performances. }
    \label{fig:tax}
\end{figure}

We compared the TaxAgent with three baseline systems and identified reasons for their sub-optimality:
\begin{itemize} 
    \item \textbf{Saez Optimal Taxation: Regressiveness}: While effective in the medium term, Saez's regressive taxation (higher rates for lower-income groups) impaired long-term performance, as shown in Figures~\ref{fig:e_q_separated} and~\ref{fig:tax}. Despite being theoretically optimal, regressive policies encounter substantial resistance in practical applications.
    \item \textbf{US Federal Income Tax: Inflexibility}: Fixed tax rates in the US system ensured stable equality (0.6–0.65) and gradual productivity growth, as illustrated in Figure~\ref{fig:e_q_separated}. However, the lack of dynamic adjustments constrained further optimization.
    \item \textbf{Free Market: Low Productivity}: The free market exhibited minimal productivity and equality, aligning with empirical findings. Without taxation and redistribution, societal productivity stagnates. This highlights the advantage of our evaluation framework over traditional models, which often assume free-market Pareto optimality under idealized conditions.
\end{itemize}

\subsubsection{Macroeconomic Side-Effects of the TaxAgent (RQ3)}
\begin{figure}
    \centering
    \includegraphics[width=1\linewidth]{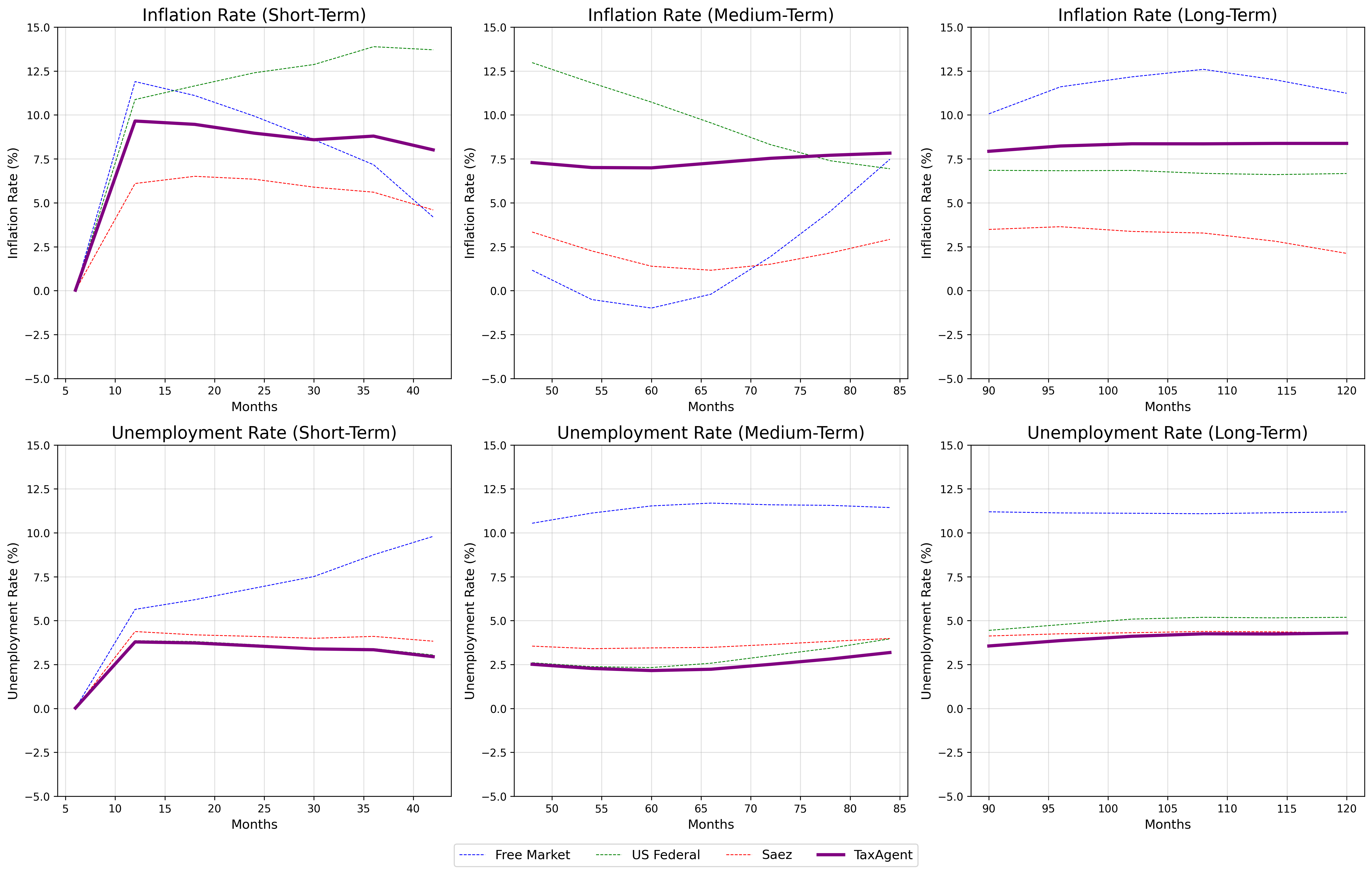}
    \caption{The side-effects of the TaxAgent on the macroeconomic environment.}
    \label{fig:side-effects}
\end{figure}

~\newline
As shown in Figure~\ref{fig:side-effects}:
\begin{itemize}
    \item \textbf{Short-term (0–40 months):} The TaxAgent performs similarly to the Saez taxation, US federal tax, and free market in controlling inflation and unemployment.
    \item \textbf{Medium-term (40–80 months):} The TaxAgent maintains stable inflation but underperforms Saez taxation, which achieves both low and stable inflation. The US federal tax and free market show inflation instability. Unemployment is stable across systems, but the free market has significantly higher rates.
    \item \textbf{Long-term (80–120 months):} Inflation and unemployment stabilize across all systems. Saez taxation has the lowest inflation, while the free market fares the worst, with the highest inflation and unemployment.
\end{itemize}

Based on the information, we conclude:
\begin{itemize}
    \item \textbf{Stable Inflation Rate:} The TaxAgent maintained a stable inflation rate of approximately 8 $\%$. Although not low, this rate is healthy compared to the high volatility observed in the free market. Nevertheless, Seaz optimal taxation is the best in inflation control.
    \item \textbf{Low Unemployment Rate:} The TaxAgent, along with Saez and US federal systems, achieved unemployment rates between 2\% to 5\%, a highly favorable outcome. The free market, however, struggles, exceeding 10\%.
\end{itemize}
These findings indicate that the TaxAgent’s superior equality-productivity balance does not come at the cost of macroeconomic instability.

\section{Conclusion}
This study introduced the TaxAgent, an innovative LLM-based tax planner, and evaluated its performance within an evaluation framework designed to emulate real-world dynamics more accurately than traditional rule-based systems. Among the four tax systems tested, the TaxAgent achieved the most favorable social outcome, surpassing even the classical Saez Optimal Taxation framework. By enhancing equality without substantially compromising productivity, the TaxAgent demonstrated a promising approach to tax planning and fiscal policy design. This work highlights a novel pathway toward creating a more equitable and prosperous society through advanced AI-driven tax strategies.

\bibliographystyle{IEEEbib}
\bibliography{main}

\begin{thebibliography}{10}

\bibitem{FLUG1998465}
Karnit Flug, Antonio Spilimbergo, and Erik Wachtenheim,
\newblock ``Investment in education: do economic volatility and credit constraints matter?,''
\newblock {\em Journal of Development Economics}, vol. 55, no. 2, pp. 465--481, 1998.

\bibitem{10.1093/isr/viab058}
Ines~A Ferreira, Rachel~M Gisselquist, and Finn Tarp,
\newblock ``On the impact of inequality on growth, human development, and governance,''
\newblock {\em International Studies Review}, vol. 24, no. 1, pp. viab058, 01 2022.

\bibitem{GLAESER2003199}
Edward Glaeser, Jose Scheinkman, and Andrei Shleifer,
\newblock ``The injustice of inequality,''
\newblock {\em Journal of Monetary Economics}, vol. 50, no. 1, pp. 199--222, 2003.

\bibitem{NBERw21340}
Hilary~W Hoynes and Ankur~J Patel,
\newblock ``Effective policy for reducing inequality? the earned income tax credit and the distribution of income,''
\newblock Working Paper 21340, National Bureau of Economic Research, July 2015.

\bibitem{NBERw21211}
Austin Nichols and Jesse Rothstein,
\newblock ``The earned income tax credit (eitc),''
\newblock Working Paper 21211, National Bureau of Economic Research, May 2015.

\bibitem{10.2307/2296779}
J.~A. Mirrlees,
\newblock ``An exploration in the theory of optimum income taxation12,''
\newblock {\em The Review of Economic Studies}, vol. 38, no. 2, pp. 175--208, 04 1971.

\bibitem{RePEc:aea:aecrev:v:61:y:1971:i:1:p:8-27}
Peter Diamond and James Mirrlees,
\newblock ``Optimal taxation and public production: I--production efficiency,''
\newblock {\em American Economic Review}, vol. 61, no. 1, pp. 8--27, 1971.

\bibitem{ATKINSON197655}
A.B. Atkinson and J.E. Stiglitz,
\newblock ``The design of tax structure: Direct versus indirect taxation,''
\newblock {\em Journal of Public Economics}, vol. 6, no. 1, pp. 55--75, 1976.

\bibitem{10.1111/1467-937X.00166}
Emmanuel Saez,
\newblock ``Using elasticities to derive optimal income tax rates,''
\newblock {\em The Review of Economic Studies}, vol. 68, no. 1, pp. 205--229, 01 2001.

\bibitem{li2024econagentlargelanguagemodelempowered}
Nian Li, Chen Gao, Mingyu Li, Yong Li, and Qingmin Liao,
\newblock ``Econagent: Large language model-empowered agents for simulating macroeconomic activities,'' 2024.

\bibitem{Foo2019ProcessAC}
Check~Woo Foo,
\newblock ``Process and critical approaches to solving the systemic climate change governance problem,''
\newblock {\em Politics \& Energy eJournal}, 2019.

\bibitem{Patjoshi2015DesignAD}
Rajesh~Kumar Patjoshi,
\newblock ``Design and development of advanced control strategies for power quality enhancement at distribution level,''
\newblock 2015.

\bibitem{10.1257/jep.25.4.165}
Peter Diamond and Emmanuel Saez,
\newblock ``The case for a progressive tax: From basic research to policy recommendations,''
\newblock {\em Journal of Economic Perspectives}, vol. 25, no. 4, pp. 165–90, December 2011.

\bibitem{10.1257/pol.6.1.230}
Thomas Piketty, Emmanuel Saez, and Stefanie Stantcheva,
\newblock ``Optimal taxation of top labor incomes: A tale of three elasticities,''
\newblock {\em American Economic Journal: Economic Policy}, vol. 6, no. 1, pp. 230–71, February 2014.

\bibitem{10.1257/pol.20180033}
Kory Kroft, Kavan Kucko, Etienne Lehmann, and Johannes Schmieder,
\newblock ``Optimal income taxation with unemployment and wage responses: A sufficient statistics approach,''
\newblock {\em American Economic Journal: Economic Policy}, vol. 12, no. 1, pp. 254–92, February 2020.

\bibitem{zheng2020aieconomistimprovingequality}
Stephan Zheng, Alexander Trott, Sunil Srinivasa, Nikhil Naik, Melvin Gruesbeck, David~C. Parkes, and Richard Socher,
\newblock ``The ai economist: Improving equality and productivity with ai-driven tax policies,'' 2020.

\bibitem{NBERc14009}
Susan Athey,
\newblock {\em The Impact of Machine Learning on Economics}, pp. 507--547,
\newblock University of Chicago Press, January 2018.

\bibitem{AxtellFarmer2022}
R.~Axtell and J.~Farmer,
\newblock ``Agent based modeling in economics and finance: Past, present, and future,''
\newblock {\em Journal of Economic Literature}, 2022.

\bibitem{DelliGatti2018}
Domenico Delli~Gatti, Giorgio Fagiolo, Mauro Gallegati, Matteo Richiardi, and Alberto Russo,
\newblock {\em Contents}, p. vii–v,
\newblock Cambridge University Press, 2018.

\bibitem{shen2025phyxdoesmodelwits}
Hui Shen, Taiqiang Wu, Qi~Han, Yunta Hsieh, Jizhou Wang, Yuyue Zhang, Yuxin Cheng, Zijian Hao, Yuansheng Ni, Xin Wang, Zhongwei Wan, Kai Zhang, Wendong Xu, Jing Xiong, Ping Luo, Wenhu Chen, Chaofan Tao, Zhuoqing Mao, and Ngai Wong,
\newblock ``Phyx: Does your model have the "wits" for physical reasoning?,'' 2025.

\bibitem{zhao2024competeaiunderstandingcompetitiondynamics}
Qinlin Zhao, Jindong Wang, Yixuan Zhang, Yiqiao Jin, Kaijie Zhu, Hao Chen, and Xing Xie,
\newblock ``Competeai: Understanding the competition dynamics in large language model-based agents,'' 2024.

\bibitem{nie2024surveylargelanguagemodels}
Yuqi Nie, Yaxuan Kong, Xiaowen Dong, John~M. Mulvey, H.~Vincent Poor, Qingsong Wen, and Stefan Zohren,
\newblock ``A survey of large language models for financial applications: Progress, prospects and challenges,'' 2024.

\bibitem{dawid2018agent}
Herbert Dawid and Domenico~Delli Gatti,
\newblock ``Agent-based macroeconomics,''
\newblock {\em Handbook of computational economics}, vol. 4, pp. 63--156, 2018.

\end{thebibliography}

\appendix
\paragraph{An example of a complete prompt of a H-Agent}

\begin{mdframed}[backgroundcolor=gray!20, skipabove=\baselineskip, skipbelow=\baselineskip]
You're Adam Mills, a 58-year-old individual living in San Antonio, Texas. A tax planner adjusts your tax rates periodically. Now it's 2001.03. Last month, you worked as a(an) Newspaper Delivery. If you continue working this month, your expected income will be \$567.18, which decreased compared to last month due to deflation of the labor market. Besides, your consumption was \$544.68. Part of your income last month was witheld as income tax. Last month, the tax brackets are: [0.00, 808.33, 3289.58, 7016.67, 13393.75, 17008.33, 42525.00] and their corresponding rates are: [0.10, 0.12, 0.22, 0.24, 0.32, 0.35, 0.37]. Income earned within each bracket is taxed only at that bracket's rate. This month, according to the tax planner, the brackets are not changed. But the planner updated corresponding rates: [10.00\%, 12.00\%, 22.00\%, 24.00\%, 32.00\%, 35.00\%, 37.00\%]. Income earned within each bracket is taxed at that bracket's rate. Pay attention to the tax rates because they may be different from the previous ones and you need to make your decision based on the current rates Deflation has led to a price decrease in the consumption market, with the average price of essential goods now at \$126.78. Your current savings account balance is \$13072.25. Interest rates, as set by your bank, stand at 3.00\%. Considering aspects like your living costs, future aspirations, broader economic trends, and the tax you need to pay, how is your willingness to work this month? How would you plan your expenditures on essential goods? Provide your decisions in a JSON format. The format should have two keys: 'work' (a value between 0 and 1 with intervals of 0.02, indicating the willingness or propensity to work) and 'consumption' (a value between 0 and 1 with intervals of 0.02, indicating the proportion of all your savings and income you intend to spend on essential goods). Keep in mind, only provide your decisions in a JSON format with two keys and two values. Do not contain any other content in your response. Keep the thinking process to yourself. I only need two key-value pairs.
\end{mdframed}

~\newline
\paragraph{An example of a complete prompt of a TaxAgent}
\begin{mdframed}[backgroundcolor=gray!20, skipabove=\baselineskip, skipbelow=\baselineskip]
You are a tax planner in charge of adjusting the tax rates of each income brackets. You will decide the tax rate in next period applied cumulatively to the income of agents in the seven [0.00, 808.33, 3289.58, 7016.67, 13393.75, 17008.33, 42525.00] income brackets. Last month, the incomes and wealth of individuals living in your society were \$[529.42, 820.63, 1255.18, 1573.74, 2015.35, 2497.29, 2781.53, 3182.72, 3753.48, 4085.31, 4744.9, 5121.11, 0.0, 0.0, 6502.85, 0.0, 7462.73, 8700.02, 0.0, 9586.97, 10203.22, 10559.54, 11607.79, 11397.85, 13069.23, 0.0, 14550.18, 15094.84, 15886.01, 16948.61, 17586.81, 20087.35, 21272.95, 21904.55, 23481.87, 24732.86, 25935.55, 27668.08, 28865.96, 30962.39, 33650.85, 35694.61, 38341.26, 40840.87, 47596.71, 52930.18, 56333.37, 0.0, 80016.97, 0.0] and \$[29273.73, 35603.8, 36976.86, 34864.24, 37601.09, 39260.21, 44126.5, 46115.94, 45439.86, 40589.7, 46877.73, 51545.67, 44164.11, 49950.69, 56873.55, 54630.91, 54940.49, 65745.22, 60696.02, 68348.58, 70040.68, 70915.03, 68205.31, 73643.45, 71548.73, 73615.63, 87073.28, 87484.31, 81080.69, 96354.78, 86864.06, 106335.95, 107013.5, 81543.38, 110052.02, 120822.74, 122489.7, 96458.05, 133070.21, 139835.22, 146328.37, 150268.78, 163927.42, 176957.88, 185924.61, 170873.53, 220595.1, 180063.13, 286498.22, 294174.67]. The tax rates you set in the past months were [([[0.1, 0.12, 0.22, 0.24, 0.32, 0.35, 0.37], [0.1, 0.12, 0.22, 0.24, 0.32, 0.35, 0.37], [0.1, 0.12, 0.22, 0.24, 0.32, 0.35, 0.37], [0.1, 0.15, 0.25, 0.3, 0.35, 0.4, 0.45], [0.1, 0.15, 0.25, 0.3, 0.35, 0.4, 0.45], [0.1, 0.15, 0.25, 0.3, 0.35, 0.4, 0.45]])]. The average per-capita productivity in the last months were [ (0.0),  (11.26),  (14.32),  (15.13),  (15.28),  (14.54)]: the past months' equality performances were [ (0.0),  (0.66),  (0.66),  (0.66),  (0.67),  (0.67)](the higher, the more equal). Adjust the tax rates to build a society that you consider best for society. You have the total freedom to adjust the rates! Provide your decision in a JSON format. The decision should be a list with seven values (each value between 0 and 1 with intervals of 0.01). Please only provide me a list with seven values between 0 and 1! Do not provide anything else! Keep the thinking process to yourself.
\end{mdframed}

~\newline
\paragraph{Calculations of Economic Metrics in the Simulation Environment}

The production is determined by the total labor supplied by households. For simplicity, we assume the production of a single homogeneous commodity, with each household contributing 168 hours (21 eight-hour working days) of labor if employed in a given period. The total production, S, is defined as:
\begin{equation}
S = \sum_{j=1}^{N} l_j \times 168 \times A
\end{equation}

Where $l_j$ represents the labor supplied by household $j$, and $A$ denotes productivity. The cumulative inventory G is updated after production as follows:

Taxation is modeled using a progressive, bracketed structure. The tax levied on a household with income $z_i$ is given by:

\begin{align}
    T(z_i) &= \sum_{k=1}^{B} \tau_{k}\left(\left(b_{k+1}-b_{k}\right) \mathbf{1}\left[z_i>b_{k+1}\right]\right. \nonumber \\
    &\left.+\left(z_i-b_{k}\right) \mathbf{1}\left[b_{k}<z_i \leq b_{k+1}\right]\right),
\end{align}

Redistribution in the simulation is even and latent. The actual post-tax income for a household is:
\begin{equation}
    {z}_i = z_i^{pre} - T(z_i) + z^r = z_i - T(z_i) + \frac{1}{N}\sum_{j=1}^N T(z_j),
\end{equation}

Demand for commodities is inversely proportional to price and directly proportional to wealth. Total societal demand is expressed as:
\begin{equation}
    D = \sum_{j=1}^N d_j = \sum_{j=1}^N \frac{c_j}{P} = \sum_{j=1}^N \frac{p_j^c s_j}{P},
\end{equation}
where $c_j$ stands for individual consumption intention; $p_c^j$ is the working propensity and $s_j$ is the accumulated wealth.

Due to inventory constraints, actual consumption is bounded by available supply:
\begin{equation}
    \hat{d}_j = \min(d_j, G), \hat{c}_j = \hat{d}_j\times P
\end{equation}
To ensure fairness, households consume in a randomized sequence, with the inventory updated after each transaction:
\begin{equation}
    G \leftarrow G - \hat{d}_j.
\end{equation}

Interest rate is defined by the Taylor Rule:
\begin{equation}
    r = \max(r^n + \pi^t + \alpha^\pi(\pi - \pi^t) + \alpha^u(u^n - u), 0),
\end{equation}
where, $r_n$  represents the natural interest rate, $\pi^t$ is current inflation, and $u$ is the unemployment rate. 

Demand-supply mismatch is quantified as:
\begin{equation}
    \Bar{\varphi} = \frac{D - G}{\max(D, G)},
\end{equation}
This imbalance triggers price and wage adjustments modeled as:
\begin{equation}
    w_i \leftarrow w_i (1 + \varphi_i), \varphi_i \sim sign(\Bar{\varphi}) U(0, \alpha_w\vert\Bar{\varphi}\vert),
\end{equation}
\begin{equation}
    P \leftarrow P (1 + \varphi_P), \varphi_P \sim sign(\Bar{\varphi})U(0, \alpha_P\vert\Bar{\varphi}\vert)
\end{equation}
where, $\alpha_p$ and $\alpha_w$ represents the maximum adjusting rates of prices and wages, respectively.

Inflation is defined as:
\begin{equation}
    \pi =\dfrac{\overline{p}_{n}-\overline{p}_{n-1}}{\overline{P}_{n-1}}
\end{equation}

Unemployment is defined as:
\begin{equation}
u = \frac{\sum_{m=1}^{12} \sum_{j=1}^{N} (1 - l_j)}{12N}
\end{equation}

Equality is defined as:
\begin{equation}
\text{eq}(x_c) = \left( 1 - \text{gini}(x_c) \right) \times \frac{N - 1}{N}
\end{equation}
where $\text{gini}(x_c)$ is the standard Gini Index of the wealth of H-Agents.

Productivity is defined as:
\begin{equation}
\text{prod}(x_c) = \sum_{i=1}^{N} x_c^i
\end{equation}

\paragraph{US Federal Income Tax}

The United States federal income tax system operates under a progressive tax structure, whereby individuals and households are taxed at increasing rates as their taxable income rises. This framework is designed to ensure that taxpayers with higher earnings contribute a proportionally larger share of their income to fund federal programs and services. The tax brackets and corresponding rates for the 2024 tax year are as follows:

\begin{itemize}
    \item 12\% Tax Rate: Applied to taxable income over \$11,000 but not exceeding \$44,725 for single filers, and over \$22,000 but not exceeding \$89,450 for married couples filing jointly.
    \item 22\% Tax Rate: Applied to taxable income over \$44,725 but not exceeding \$95,375 for single filers, and over \$89,450 but not exceeding \$190,750 for married couples filing jointly.
    \item 24\% Tax Rate: Applied to taxable income over \$95,375 but not exceeding \$182,100 for single filers, and over \$190,750 but not exceeding \$364,200 for married couples filing jointly.
    \item 32\% Tax Rate: Applied to taxable income over \$182,100 but not exceeding \$231,250 for single filers, and over \$364,200 but not exceeding \$462,500 for married couples filing jointly.
    \item 35\% Tax Rate: Applied to taxable income over \$231,250 but not exceeding \$578,125 for single filers, and over \$462,500 but not exceeding \$693,750 for married couples filing jointly.
    \item 37\% Tax Rate: Applied to taxable income exceeding \$578,125 for single filers and \$693,750 for married couples filing jointly.
\end{itemize}

\paragraph{The Saez Optimal Taxation}
The Saez tax framework is formalized as hereunder.

The utility of an individual depends positively on consumption \( c \) and negatively on labor effort \( z \), and is given by:
\begin{equation}
u(c,z) = v(c) - h(z),
\end{equation}
where \( v(c) \) captures the utility from consumption, and \( h(z) \) represents the disutility from labor effort. Individuals face a budget constraint:
\begin{equation}
c = z(1-\tau) + R,
\end{equation}
where \( z \) is earnings, \( \tau \) is the marginal tax rate, and \( R \) is virtual income.

Behavioral responses to taxation are captured through three key elasticities:
\begin{itemize}
    \item \textbf{Uncompensated Elasticity} (\( \epsilon_u \)): 
    \begin{equation}
    \epsilon_u = \frac{1-\tau}{z} \cdot \frac{\partial z}{\partial(1-\tau)},
    \end{equation}
    which measures the sensitivity of earnings to changes in the net-of-tax rate \( (1-\tau) \);
    \item \textbf{Income Effect} (\( \eta \)): 
    \begin{equation}
    \eta = \frac{1-\tau}{z} \cdot \frac{\partial z}{\partial R},
    \end{equation}
    which represents how changes in virtual income influence labor supply;
    \item \textbf{Compensated Elasticity} (\( \epsilon_c \)): 
    \begin{equation}
    \epsilon_c = \epsilon_u + \eta,
    \end{equation}
    which captures the pure substitution effect after accounting for income effects.
\end{itemize}

The government’s objective is to maximize social welfare:
\begin{equation}
W = \int_z w(z) u(c,z) \, dz,
\end{equation}
where \( w(z) \) are welfare weights, decreasing with income to reflect redistributive goals.

For high-income earners, Saez derives a simple formula for the optimal marginal tax rate:
\begin{equation}
\tau^* = \frac{1}{1 + a \cdot \epsilon_u},
\end{equation}
where \( a = \frac{\bar{z}}{\bar{z} - z^*} \) is the Pareto parameter, reflecting the thickness of the income distribution's top tail. This formula balances revenue gains from increased tax rates with losses due to reduced labor supply, ensuring progressivity without excessive distortion.

Extending to the full income distribution, Saez provides a general nonlinear tax schedule:
\begin{equation}
T'(z) = \frac{(1 - G(z)) + e \cdot z \cdot g(z)}{1 + e \cdot g(z)},
\end{equation}
where \( G(z) \) is the cumulative income distribution, \( g(z) \) is the income density, and \( e \) is the elasticity of taxable income.

Saez’s framework emphasizes progressive taxation with higher marginal rates for top earners, justified by diminishing marginal utility of income and empirical evidence on elasticities. 

\paragraph{Ablation Study}

\begin{figure}
    \centering
    \includegraphics[width=1\linewidth]{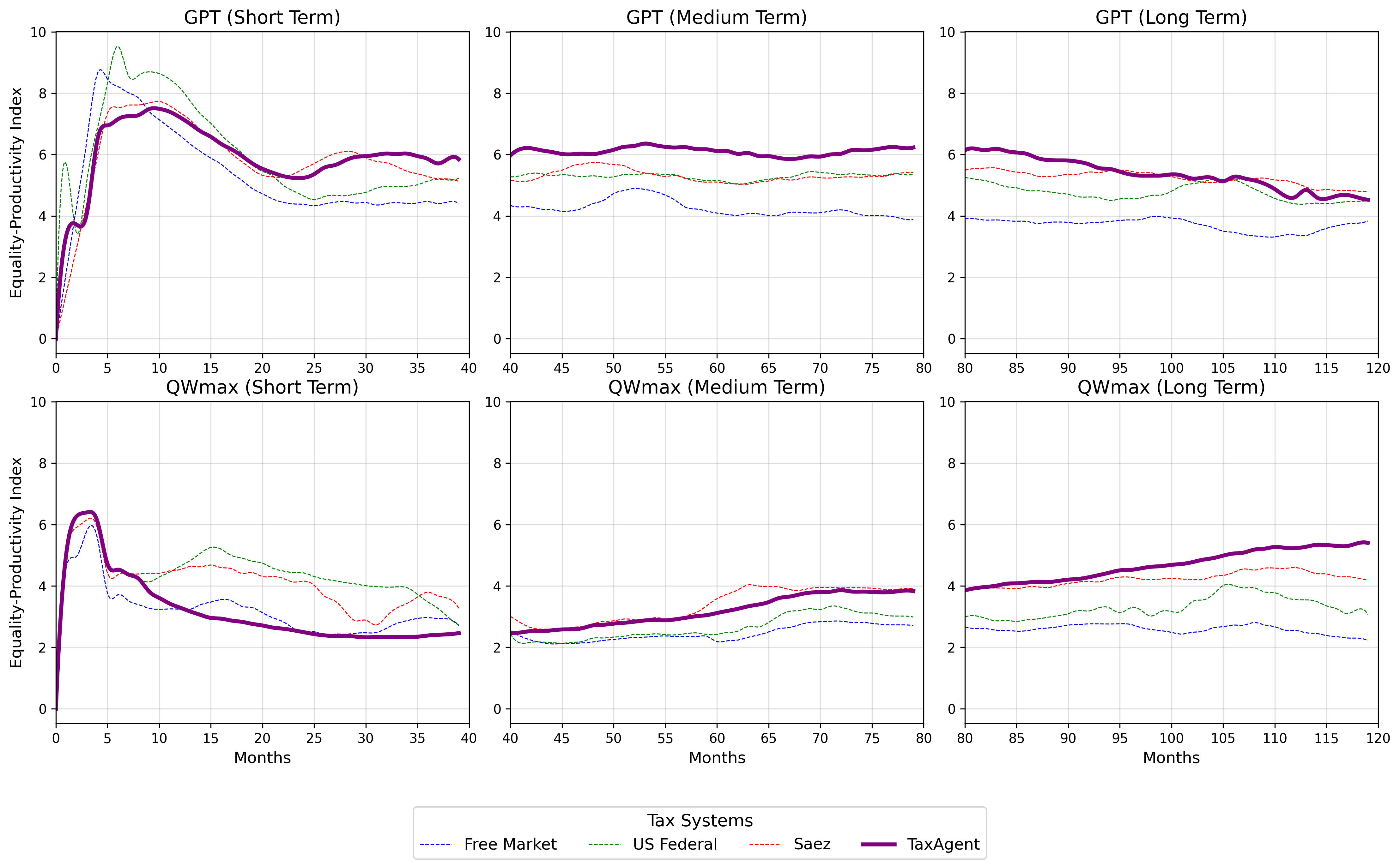}
    \caption{Ablation study of the robustness of the TaxAgent. The TaxAgent shows low sensitivity to changes in its base LLM. }
    \label{fig:ablation}
\end{figure}

The results of using qwen-max-2024-09-19 and gpt-4o-2024-08-06 as the TaxAgent are shown in Figure~\ref{fig:ablation}. In general, the social outcome generated by the TaxAgent is superior in the long term. An exception is that its performance experiences a slight drop after the 100th month when the base LLM is Chatgpt, but the performance is not significantly lower than its competitors'. This indicates that the TaxAgent has low sensitivity to changes in its LLM base, enhancing its reliability.

\vspace{12pt}

\end{document}